\documentclass[runningheads]{llncs}

\usepackage{eccv}

\usepackage{eccvabbrv}

\usepackage{graphicx}
\usepackage{booktabs}

\usepackage[accsupp]{axessibility}  

\usepackage{multirow}
\usepackage{colortbl}
\usepackage{xcolor}
\newcolumntype{G}{>{\color{gray!50}}c}

\usepackage{multicol}
\usepackage{makecell}

\usepackage[pagebackref,breaklinks,colorlinks,citecolor=eccvblue]{hyperref}

\usepackage{orcidlink}

\begin{document}

\newcommand{\name}{$\mathtt{RealTalk}$}

\title{RealTalk: Real-time and Realistic Audio-driven Face Generation with 3D Facial Prior-guided Identity Alignment Network} 

\titlerunning{RealTalk}

\author{Xiaozhong Ji \inst{1} \and
Chuming Lin \inst{1} \and
Zhonggan Ding\inst{1} \and
Ying Tai \inst{2} \and
Junwei Zhu \inst{1} \and
Xiaobin Hu \inst{1} \and
Donghao Luo \inst{1} \and
Yanhao Ge \inst{3} \and
Chengjie Wang \inst{1}
}

\authorrunning{Ji et al.}

\institute{Youtu Lab, Tencent \\
\email{xiaozhongji@tencent.com}\\
\and Nanjing University\\
\email{yingtai@nju.edu.cn}\\
\and VIVO\\
\email{halege@vivo.com}}

\maketitle


\begin{abstract}

Person-generic audio-driven face generation is a challenging task in computer vision.
Previous methods have achieved remarkable progress in audio-visual synchronization, but there is still a significant gap between current results and practical applications.
The challenges are two-fold: 
$1$) Preserving unique individual traits for achieving high-precision lip synchronization.
$2$) Generating high-quality facial renderings in real-time performance.
In this paper, we propose a novel generalized audio-driven framework \name, which consists of an audio-to-expression transformer and a high-fidelity expression-to-face renderer.
In the first component, we consider both identity and intra-personal variation features related to speaking lip movements. 
By incorporating cross-modal attention on the enriched facial priors, we can effectively align lip movements with audio, thus attaining greater precision in expression prediction.
In the second component, we design a lightweight facial identity alignment (FIA) module which includes a lip-shape control structure and a face texture reference structure. 
This novel design allows us to generate fine details in real-time, without depending on sophisticated and inefficient feature alignment modules.
Our experimental results, both quantitative and qualitative, on public datasets demonstrate the clear advantages of our method in terms of lip-speech synchronization and generation quality.
Furthermore, our method is efficient and requires fewer computational resources, making it well-suited to meet the needs of practical applications.
\end{abstract}
    
\vspace{-5mm}
\section{Introduction}
\label{sec:intro}

\begin{figure*}[ht!]
    \centering
	\captionsetup{type=figure}
	\includegraphics[width=1\linewidth]{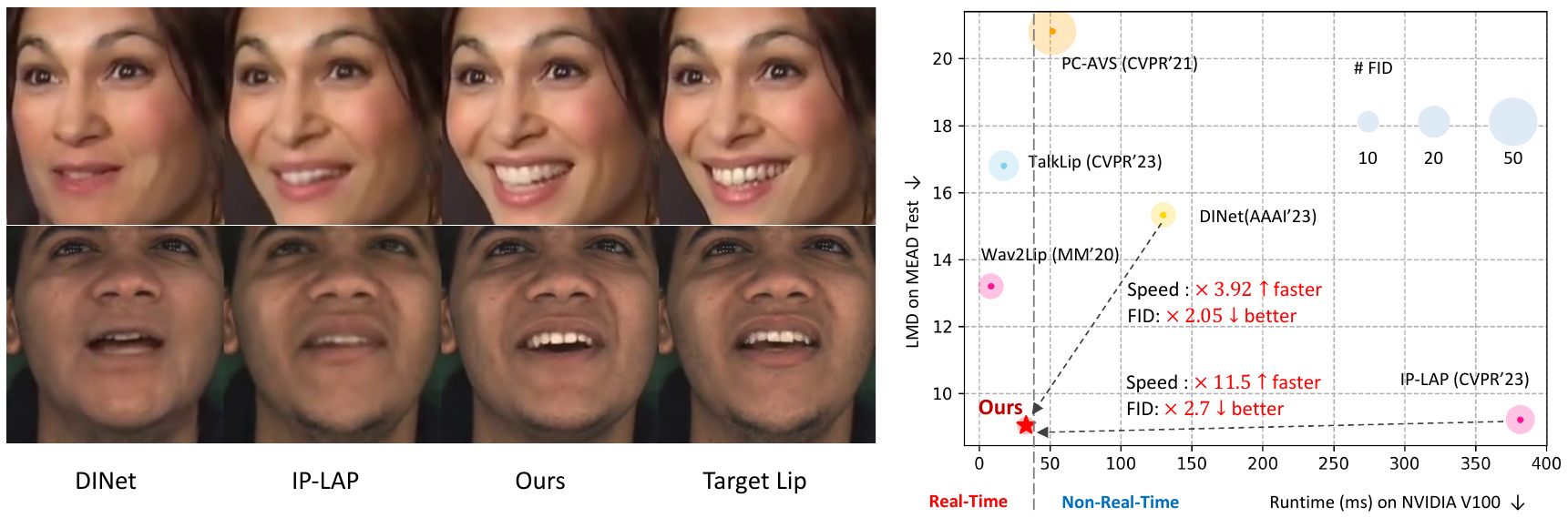} 
	\vspace{-6mm}
	\captionof{figure}{\textbf{Left: Visual Comparison on lip sync and generation quality with IP-LAP~\cite{zhong2023identity} and DINet~\cite{zhang2023dinet}.} Our method achieves precise lip-synced talking faces, closer to the target lip, with higher visual quality.  \textbf{Right: Speed, LMD and FID comparisons}. Our method generates talking faces at $30$ FPS on NVIDIA V$100$, showcasing the \textit{best} LMD and FID scores while maintaining the \textit{real-time} speed.}
	\label{fig:shortshow}
\vspace{-2mm}

\end{figure*}

Audio-driven face generation has received much attention in recent years due to its great potential in real-world applications. 
The main challenges in generating realistic and expressive talking faces include:
$1$) Ensuring lip-speech synchronization that matches the audio input and the lip movements.
$2$) Achieving photo-realistic visual quality that preserves the details and textures of the face.
$3$) Maintaining identity preservation that keeps the expressions and facial features consistent with the original individual.
$4$) Enhancing efficiency that enables fast and robust face generation, especially for applications in real-world scenarios.

Existing person-generic talking face generation methods can be roughly divided into two categories: realtime-based and non realtime-based methods as listed in Table~\ref{tab:method_comparsion}.
In the first group, Wav2Lip~\cite{prajwal2020lip} employs a sync-expert to improve the lip-synchronization performance with accurate lip motion.
TalkLip~\cite{wang2023seeing} proposes an efficient framework that tackles the reading intelligibility problem by leveraging a lipreading expert.
These methods \textit{have fast inference speed but usually suffer from the unsatisfactory generation effects}, \textit{e.g.} blurry faces in Wav2Lip or facial artifacts in TalkLip. 

For methods in the second category,
PC-AVS~\cite{zhou2021pose} incorporates disentangle learning for identity, speech content, and poses in talking face generation.
DINet~\cite{zhang2023dinet} develops a deformation part and an inpainting part for accurate mouth movements and textual details. 
StyleTalk~\cite{ma2023styletalk} utilize an implicit style code to control global head and facial movements. 
IP-LAP~\cite{zhong2023identity} utilizes the guidance of prior landmark and appearance information, and proposes a two-stage framework, consisting of an audio-to-landmark generator and a landmark-to-video generation model.
IP-LAP \textit{produces better visual results than the realtime methods but is time-consuming}, making it impractical for real-world applications.

\begin{table}[ht!]
    \centering

    \caption{\textbf{Method categories and complexity comparsions}. 
    The primary distinction between~\name~and existing methods lies in the incorporation of cross-modal attention on $3$D priors, learnable mask and the FIA module. 
    The methods highlighted in \textbf{bold} are capable of real-time performance. 
    Furthermore, our method attains real-time performance via its compact structure and reduced dependency on reference frames.}
    \vspace{-3mm}
    \label{tab:method_comparsion}
    \renewcommand{\arraystretch}{1.0}
    \setlength\tabcolsep{3.0pt}
    \resizebox{0.9\linewidth}{!}{

\setlength{\tabcolsep}{8pt}

\begin{tabular}{@{}llllc@{}}
        \toprule
        \multirow{3}{*}{\makecell[c]{Method\\ \vs \\ Category}} & \multirow{3}{*}{\makecell[c]{Audio to Face \\ Translation}} & \multirow{3}{*}{\makecell[c]{Facial \\ Mask}} & \multirow{3}{*}{\makecell[c]{Identity Alignment \\ ($n$ frames)}} & \multirow{3}{*}{\makecell[c]{Speed \\(FPS)}} \\
        & & & & \\
        & & & & \\
        \toprule
        \textbf{Wav2Lip}~\cite{prajwal2020lip}   & Audio encoder & Lower-half  & Concat ($1$)       & $120$ \\
        \textbf{TalkLip}~\cite{wang2023seeing}   & Audio encoder & Lower-half  & Concat ($1$)       & $57$ \\
        PC-AVS~\cite{zhou2021pose}      & Audio encoder & Full face & Concat (1)       & $17$ \\
        DINet~\cite{zhang2023dinet}     & Audio encoder & Rectangle  & Deformation ($5$)   & $8$ \\
        StyleTalk~\cite{ma2023styletalk} & Style decoder   & Full face & Flow-based ($1$)    & $7$ \\ 
        IP-LAP~\cite{zhong2023identity} & landmark transformer  & Lower-half  & Flow-based ($25$)   & $3$ \\ \midrule
        \textbf{RealTalk (Ours)}                 & \makecell[l]{3D cross-modal\\temporal attention}  & Learnable   & FIA module ($1$)    & $30$ \\
        \bottomrule
        \end{tabular}
    }
\end{table}

To achieve realtime efficiency and high-fidelity talking face effects simultaneously, in this paper we propose a novel framework, termed \name, which consists of an audio-to-expression transformer converting input audio into $3$D expression coefficients, and an expression-to-face renderer generating high-fidelity talking face from the estimated $3$D expression.
Specifically, there are three key designs in \name~to improve the performance and efficiency:

\noindent $1$) \textit{Improved facial prior with cross-modal attention} in audio-to-expression transformer. 
Previous work~\cite{chen2012jointbayesian} observed and discussed that the appearance of a face is influenced by two factors: identity and intra-personal variation (\textit{e.g.}, expression, pose, lighting).
Inspired by this observation, we enrich the input facial prior by performing cross-modal on the $3$D shape and historical expression coefficients as $3$D facial prior guidance besides the input audio queries. Here, the shape represents the identity, while expressions from historical frames capture intra-individual lip amplitude variations.

\noindent $2$) \textit{Learnable facial mask} as the bridge connecting the two networks.
Different from the previous methods that occlude half of the face or adopt a fixed position black square, our method leverages the learned $3$D expressions from the audio-to-expression transformer, and converts them into an adaptive facial mask that better estimates the output facial structure given the input audio, leading to better performance in facial contour generation and lip motion accuracy. 

\noindent $3$) \textit{Efficient and effective network design} in expression-to-face renderer.
We highlight the advantages of our FIA module in inference speed, which dominates the overall runtime. 
Unlike recent methods~\cite{zhong2023identity,zhang2023dinet} that require \textit{time-consuming} feature extraction from multiple reference images to enhance visual quality,
our FIA module is meticulously crafted to achieve high-quality texture generation in \textit{real-time} using only $1$ image.
Specifically, different from the methods that use optical flow~\cite{zhong2023identity} or deformation module~\cite{zhang2023dinet}, our method designs a novel Facial Identity Alignment (FIA) module to achieve high-fidelity talking face synthesis.

Overall, our contributions are summarized as follows:
\begin{itemize}

\item The proposed \name~makes full use of the improved 3D facial prior by applying cross-modal attention to shape and variation to help predict more accurate facial expressions.

\item The proposed FIA module exhibits strong control over lip movements and texture referencing capabilities, thereby producing high-quality facial images without sacrificing efficiency.

\item To our best knowledge, the proposed \name~is the best choice considering both accuracy and efficiency (\textit{i.e.}, $30$FPS) for talking face generation as shown in Fig.~\ref{fig:shortshow}.

\end{itemize}

\section{Related Work}
\label{sec:related_works}



\noindent \textbf{Audio-driven Talking Face Generation.}
Existing audio-driven face generation can be primarily divided into two categories of methods: person-specific and person-generic approaches. Person-specific methods ~\cite{thies2020neural, zhang2021facial,lu2021live,guo2021ad, liu2022semantic, shen2022learning,ye2023geneface} require training or fine-tuning on specific individuals before inference, whereas person-generic methods~\cite{chen2019hierarchical,prajwal2020lip,hu2020face,zhou2021pose,cheng2022videoretalking,guan2023stylesync,ji2022eamm,ma2023styletalk,park2022synctalkface,shen2023difftalk,sun2022masked,wang2023seeing,xu2023emotiontalk,zhang2023dinet,zhong2023identity,zhu2022hifihead} enable the direct generation of talking face videos for unseen person. To address the challenge of audio-visual synchronization in person-generic methods, Wav2Lip~\cite{prajwal2020lip} introduces a lip synchronization discriminator using SyncNet~\cite{chung2017out}. In contrast, TalkLip~\cite{wang2023seeing} utilizes a lip reading network to enhance the comprehensibility of the lip region. Furthermore, several  approaches~\cite{chen2019hierarchical,ji2022eamm,zhong2023identity,ma2023styletalk} focus on modeling the mapping from audio to facial expressions, which simplifies the process of lip synchronization. To enhance the visual quality, DiffTalk~\cite{shen2023difftalk} employs a diffusion model and StyleSync~\cite{guan2023stylesync} utilizes StyleGAN~\cite{karras2020analyzing} to provide high-fidelity facial priors. Additionally, certain methods~\cite{ji2022eamm,zhong2023identity,zhang2023dinet} employ identity reference alignment to preserve facial identity and texture. However, achieving a balance between efficiency, visual quality, and accuracy of lip movements is a formidable challenge for the aforementioned person-generic methods.
\vspace{5mm}

\noindent \textbf{Audio to Facial Expressions Modeling.}
Modeling the integration of audio into facial expressions in a general context enables more efficient and accurate learning of lip movements. 
IP-LAP~\cite{zhong2023identity} predicts facial landmarks of the target face based on the audio input, while EAMM~\cite{ji2022eamm} utilizes the unsupervised FOMM~\cite{siarohin2019first} to extract the keypoints of the target face. Although facial landmarks or keypoints are relatively easy to obtain, they suffer from sparsity, making it challenging to represent complex facial movements adequately, such as the actions of puckering or pursing the lips. To address these limitations, we propose using $3$DMM~\cite{blanz1999morphable} to extract more accurate decoupled information about facial identity, pose, and expression, and learn the mapping from audio to expression coefficients. 
Compared to $2$D facial landmarks, $3$DMM provides denser keypoints, allowing for a more comprehensive representation of intricate facial region movements.
\vspace{5mm}

\noindent \textbf{Identity Reference Alignment.}
Previous methods commonly employ encoder-decoder architectures to directly fuse reference identity frames, but they often fail to effectively preserve identity features. In contrast, identity reference alignment ensures a stronger resemblance between the generated results and the identity. For example, IP-LAP~\cite{zhong2023identity} relies on optical flow~\cite{horn1981determining} to align multiple identity reference features by warping them onto the target frame. 
DINet~\cite{zhang2023dinet} employs adaptive affine transformation~\cite{zhang2022adaptive} to process multiple reference frames and generate deformed features that enhance identity information. 
These methods employ intricate alignment strategies and \textit{multiple} identity references, slowing down the inference speed. 
In contrast, we propose an efficient facial identity alignment module by utilizing \textit{single} frame of identity reference.

\begin{figure*}[ht!]
    \centering
    \includegraphics[width=1.0\linewidth]{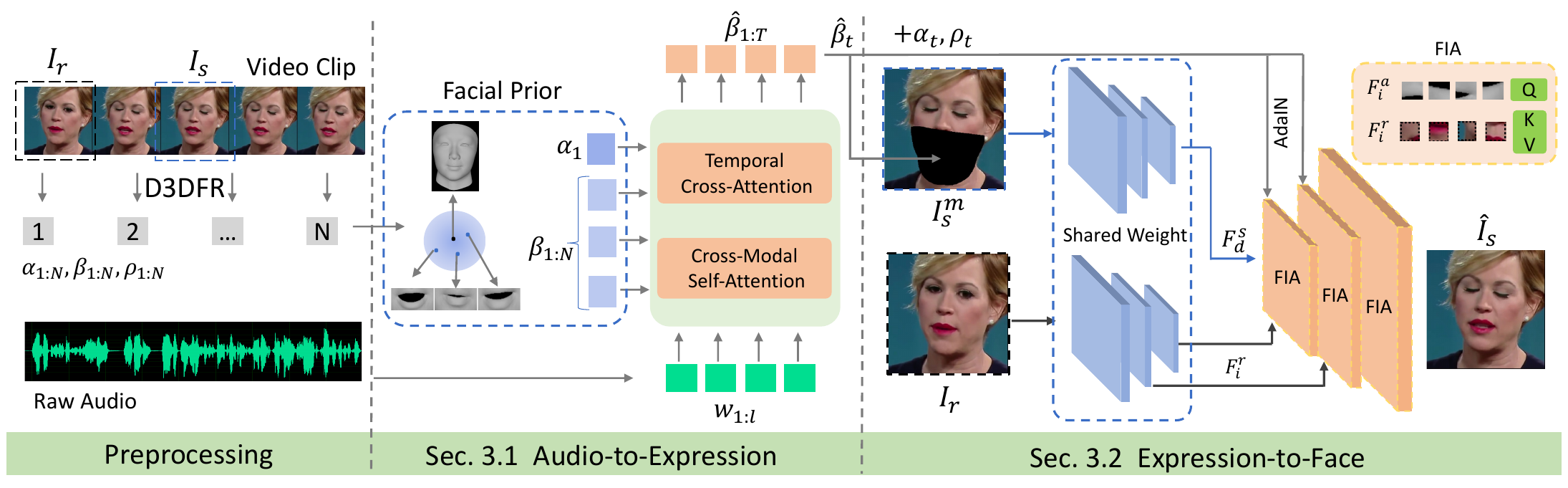} 
    \vspace{-4mm}
    \caption{\textbf{Framework of our approach.} Our network is divided into two parts: \textit{Audio-to-expression Transformer}, and \textit{Expression-to-face Renderer}. The preprocessing is to extract 3D shapes $\alpha_{1:N}$, expressions $\beta_{1:N}$, poses $\rho_{1:N}$, and audio feature $w_{1:l}$. In the first part, the shape $\alpha_{1}$ and historical expressions $\beta_{1:N}$ are utilized as \textbf{Improved Facial Prior} to predict $\hat{\beta}_{1:N}$ while preserving identity and intra-personal lip amplitude variations. In the second part, the predicted expressions are injected into the proposed \textbf{Facial Identity Alignment (FIA)} module to inpaint the masked source frame $I_s^m$ the target lip through cross-attention with the identity reference $I_r$.} 
    \label{fig:framework}
\end{figure*}

\section{Method}

\noindent \textbf{Overview.}
Our goal is to generate a video that synchronizes the lip movements with a target audio clip while maintaining the consistency of the facial identity from the original video. 
Fig.~\ref{fig:framework} illustrates our method, which consists of two stages.
In the first stage, we utilize shape and historical expressions as conditions to map the audio to $3$D expression coefficients with the proposed audio-to-expression transformer. 
In the second stage, we design an lightweight face renderer including a facial identity alignment module to generate the target lip based on the predicted expression coefficients and reference frame.

\subsection{Audio-to-expression Transformer}
\begin{figure}[t!]
	\centering
	\includegraphics[width=0.95\columnwidth]{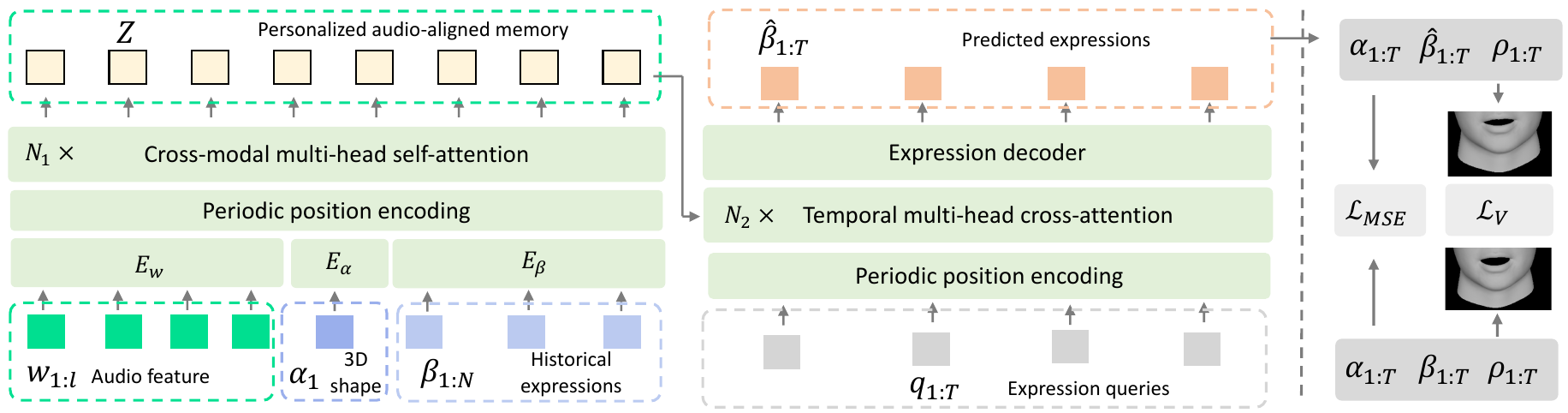} 
	\caption{\textbf{Architecture of the Audio-to-expression Transformer.} 
	(Left) The audio, shape, and historical expressions are processed through an encoder to obtain memory features, which are then combined with expression queries in decoder to generate predictions. 
	(Right) The predicted expressions and GT expressions are optimized using reconstruction and vertex losses.
	} 
	\label{fig:a2e}
\end{figure}

In this stage, our objective is to generate precise and stylistically consistent lip movements. 
Leveraging $3$D face reconstruction technology, we can effectively control dense facial regions with a reduced number of coefficients. 
Given a series of facial images,  we use the D$3$DFR model~\cite{deng2019accurate} to extract $3$D coefficients.
These coefficients consist of three components: shape $\alpha$, expression $\beta$, and pose $\rho$. 
As shown in Fig.~\ref{fig:a2e}, the audio-to-expression network takes three inputs, including driving audio features, the 3D shape feature, and historical expressions.

\noindent \textbf{Shape and Expression Prior.}
The uniqueness of each individual's facial and mouth structures results in variations in how audio and lip movements align. 
Here, we enrich two types of personalized facial priors, \textit{i.e.} shape and historical expressions. 

\textit{Shape} signifies identity, which is typically related to the natural face size and mouth proportions.
On the other hand, \textit{historical expressions} capture the individual's unique lip amplitude, which supply the individual variations from the standard shape. 


Firstly, we designate the first frame to obtain the default shape coefficient, denoted as $\alpha_1$ and $N$ frames of historical expressions, represented as $\beta_1,\cdots,\beta_N$. 
We utilize Hubert~\cite{hsu2021hubert} to extract audio features. 
The audio features can be represented as $w_1$, $\cdots$, $w_{l}$, where $l$ denotes the length of audio feature. 
Specifically, the shape, expression coefficients and audio features are passed through fully connected networks to obtain embeddings, respectively. 
These embeddings are then concatenated in sequence order, resulting in a total of $l+N+1$ tokens, which are input into a periodic position encoding layer and $N_1$ cascaded Cross-Modal multi-head Self-Attention (CMSA) to obtain the personalized audio-aligned memory:
\begin{equation}
\begin{aligned}
Z = \operatorname{CMSA}(w_1,~\cdots,w_{l}, \alpha_1, \beta_1,~\cdots, \beta_N).
\end{aligned}
\end{equation} 
The expression queries $q_t$ (initially set to zero), are combined with the memory and fed into $N_2$ cascaded  Temporal multi-head Cross-Attention (TCA) network and the linear decoder to get expression prediction:
\begin{equation}
\begin{aligned}
\hat{\beta}_1,\hat{\beta}_2,\cdots, \hat{\beta}_T = \operatorname{TCA}(q_1, q_2,~\cdots,q_{T}, Z).
\end{aligned}
\end{equation} 

\noindent \textbf{Loss Function.}
The loss function primarily consists of Mean Squared Error (MSE) and $3$D vertex loss. 
MSE calculates the error between the predicted expression coefficients and the Ground Truth (GT): 
\begin{equation}
\begin{aligned}
\mathcal{L}_{MSE} = \frac{1}{T} \Sigma^{T}_{t=1} \| \beta_t - \hat{\beta_t} \|_2^2. 
\end{aligned}
\end{equation} 
The vertex loss calculates $3$D vertices by combining the predicted expression coefficients $\hat{\beta}$ with shape $\alpha_1$ and pose $\rho$, and chooses points from the mouth region to evaluate the distance. These loss components optimize the process by minimizing the gap between the predicted and GT expression coefficients and ensure accurate alignment of the generated $3$D vertices with the mouth keypoints. 

Let V represent the $3$D vertex computed from the coefficients. The vertex loss can be described as follows:
\begin{equation}
\begin{aligned}
\mathcal{L}_{V} = \frac{1}{T} \Sigma^{T}_{t=1} \| \operatorname{V}(\alpha_t, \beta_t, \rho_t) - \operatorname{V}(\alpha_t, \hat{\beta_t}, \rho_t) \|_2^2.
\end{aligned}
\end{equation} 
The overall loss is $\mathcal{L}_{a2e} = \mathcal{L}_{MSE} + 0.1 *\mathcal{L}_{V}$. 

\subsection{Expression-to-face Renderer}

We design a lightweight network for generating facial images with edited lips. 
The face renderer takes a masked source image $I_s^m$, a reference image $I_r$, and $3$D coefficients \{$\alpha_t, \hat{\beta}_{t}, \rho_{t}$\} as inputs. 
Our network adopts an encoder-decoder architecture, where both images are processed via a shared-weight encoder to extract multi-scale features, subsequently integrated with 3D coefficients within the decoder. 
Next, we describe the detailed process below.

\begin{figure}[t!]
	\centering
	\includegraphics[width=0.7\columnwidth]{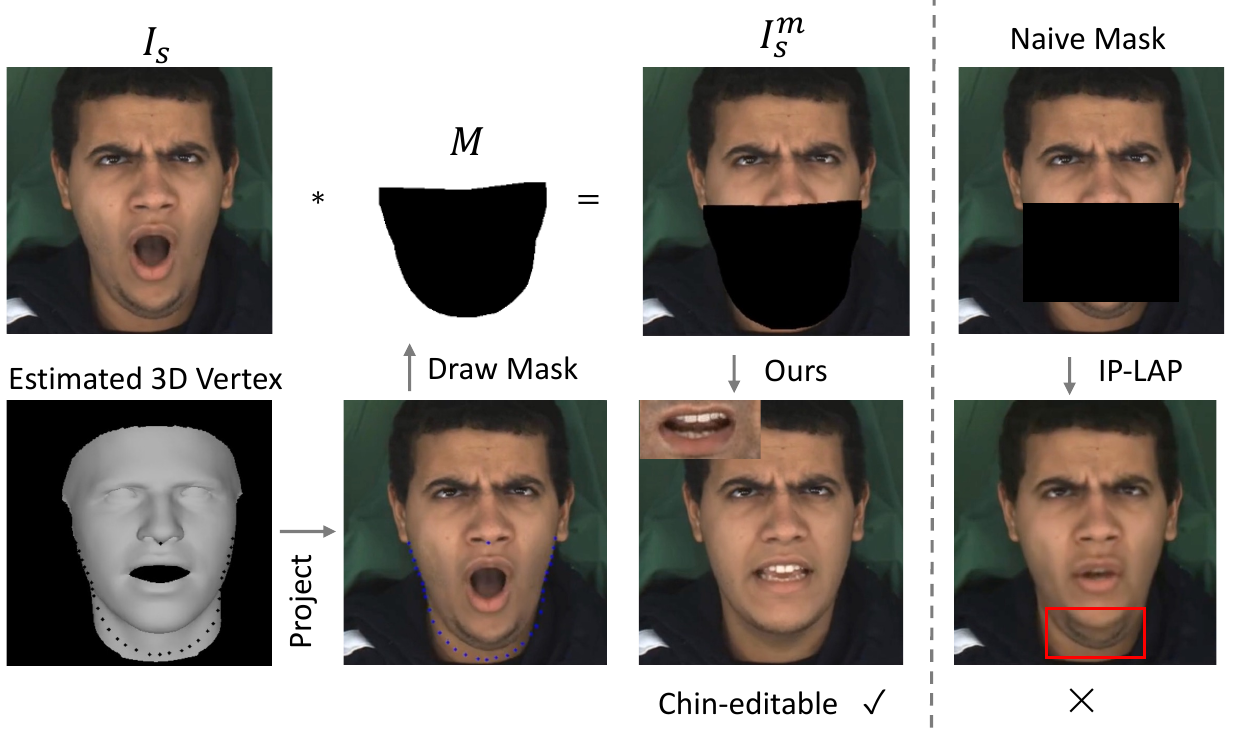} 
	\vspace{-3mm}
	\caption{\textbf{Illustration of the learnable mask} based on predicted facial expressions. 
	$1$) The estimated $3$D vertex allows us to select points with fixed positions relative to the face. We choose points that emphasizes a larger facial contour to accommodate diverse lip movements. 
	$2$) Comparisons between our learnable mask and the naive mask (IP-LAP), with the left-top in our result representing the target lip, show that our method effectively adjusts the face shape based on the spoken content, while IP-LAP yields unnatural results, \eg a double chin.
	}
	\label{fig:mask}
\vspace{-2mm}
 
\end{figure}

\noindent \textbf{Learnable Mask.}
Unlike previous methods, as shown in Fig.~\ref{fig:mask}, we selectively mask the source image $I_s$ allowing for more accurate control over the modified areas.
The $3$D vertices are estimated according to the predicted expression coefficients and then projected onto the image. 
Based on predetermined sectional points, we draw and fill the mouth and neck regions. 
The process of generating the mask is as follows:
\begin{equation}
	\begin{aligned}
		V_{xy} &=  \operatorname{P}(\operatorname{V}(\alpha_t, \hat{\beta_t}, \rho_t), \tau_t), \\
		M &= \operatorname{C}(V_{xy}), (x,y) \in S, \\ 
		I_s^m &= M * I_s,
	\end{aligned}
\end{equation} 
where $\operatorname{P}$ is the project function and $\tau_t$ indicates the translation matrix of $t$. $\operatorname{C}$ is the contour of points in the face and neck area of interest, and the set of these points is $S$.
$I_s^m$ represents the image operated with the learned mask $M$.


\begin{figure}[t!]
	\centering
	\includegraphics[width=0.7\columnwidth]{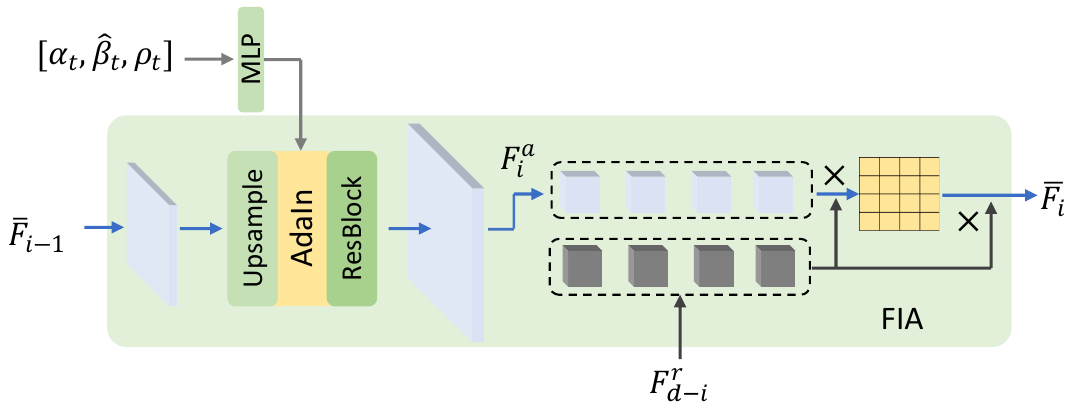} 
 	\vspace{-4mm}
	\caption{\textbf{Architecture of the Facial Identity Alignment module.} The inputs to the FIA module include the predicted facial expression coefficients $\hat{\beta}_t$, the known shape $\alpha_t$ and pose $\rho_t$, the feature from last module $\Bar{F}_{i-1}$, and the identity reference image feature $F_{d-i}^r$. The 3D coefficients are injected through AdaIN, enabling control over the lip. Multi-scale reference features interact with the current features through cross-attention, facilitating effective texture transfer.} 
	\label{fig:gen}
 \vspace{-2mm}
 
\end{figure}

\noindent \textbf{Shared Encoder.}
The encoder consists of $d$ stages of stacked residual blocks, where the resolution is reduced by half and the feature dimension is increased at each stage. 
Since the encoder weights are shared, we parallelize the computation of the source and reference along the batch dimension. 
The encoder extracts multi-scale features as:
\begin{equation}
	\begin{aligned}
		F_{i}^s, F_{i}^r = \operatorname{SE_i}([F_{i-1}^s, F_{i-1}^r]), 
	\end{aligned}
\end{equation}  
where $\operatorname{SE_i}$ represents $i$-th level of the shared encoder. Then we get $[F^s_1,~\cdots, F_d^s]$ and $[F^r_1,~\cdots, F_d^r]$.

\noindent \textbf{Facial Identity Alignment Module (FIA).}
In decoder, each layer's input features $\Bar{F}_{i-1}$, output from last stage (initially $F_{d}^s$),  are first upsampled to increase the resolution, denoted as $F_i^u$. As depicted in Fig.~\ref{fig:gen},
the $3$D coefficients \{$\alpha_t, \hat{\beta}_{t}, \rho_{t}$\} undergo dimension mapping through a three-layer MLP before being injected into the network through an AdaIN module.
The feature is modulated by the injected $3$D coefficients to match the specific lip shapes. 
We then incorporate multiple additional residual blocks ($2$ in our final model) to further enhance these features. 
Finally, features from the reference image $F_{d-i}^r$ are aligned at the same resolution to control face generation and aggregate textures: 

\begin{equation}
	\begin{aligned}
		\Bar{F}_i = \operatorname{FIA_i}([\alpha_t, \hat{\beta}_{t}, \rho_{t}],\Bar{F}_{i-1}, F_{d-i}^r).
	\end{aligned}
\end{equation}
To avoid generating unnecessary background, we adopt a blending strategy by merging the results of the final layer with the input to obtain the final outcome:
\begin{equation}
	\begin{aligned}
		\hat{I_s} = M*I_s + (1-M) *\Bar{F}_d. 
	\end{aligned}
\end{equation}

\noindent \textbf{Loss Function.}
We employ multiple loss functions to constrain the accuracy of lip-sync and visual quality, including pixel loss, perceptual loss, adversarial loss, and local pixel loss to enhance the details of teeth: 
\begin{equation}
	\begin{aligned}
		\mathcal{L}_{1} &= \Sigma \| \hat{I_s} - I_s \|_1, \\
		\mathcal{L}_{2} &= \Sigma \| \operatorname{VGG}(\hat{I_s}) - \operatorname{VGG}(I_s) \|_1, \\
		\mathcal{L}_{3} &= \mathbb{E}_{I_s}[\log{D(I_s)}+\log(1-D(\hat{I_s}))], \\
		\mathcal{L}_{4} &= \Sigma \| M'*\hat{I_s} - M'*I_s \|_1, \\
	\end{aligned}
\end{equation}
where $D(\cdot)$ is the discriminator and $M'$ represents the binary mask of the teeth area.
The overall loss function is obtained by weighting the above losses according to their respective weights. 
\begin{equation}
	\begin{aligned}
		\mathcal{L}_{e2f} &= \lambda_1 * \mathcal{L}_{1} + \lambda_2 * \mathcal{L}_{2} + \lambda_3 * \mathcal{L}_{3} + \lambda_4 * \mathcal{L}_{4},
	\end{aligned}
\end{equation} 
where $\lambda_1=1,\lambda_2=1,\lambda_3=0.1,\lambda_4=1$.

\noindent \textbf{Discussion with IP-LAP and DINet on Efficient Design.}
To reduce computational burden, we employ several optimization strategies. Firstly, we utilize highly compressed $3$D coefficients as the context for audio-to-face conversion. This approach is computationally more efficient than IP-LAP~\cite{zhong2023identity}'s use of $2$D landmarks images, thereby circumventing the need to process high-dimensional features.
Secondly, we implement a shared encoder to concurrently extract multi-scale features from both masked source and unmasked reference images.
Thirdly, we employ only $1$ frame for texture transfer, in contrast to the $5$ and $25$ frames used by DINet~\cite{zhang2023dinet} and IP-LAP respectively. For example, IP-LAP increases the computational load of the alignment module by warping each reference frame to the current image via optical flow. Conversely, DINet accomplishes mouth inpainting by extracting deformation features from reference frames.
Lastly, our FIA module is designed for efficiency. Its cross-attention mechanism can adaptively query similar texture features and can be flexibly embedded at various scales within the network. It should be noted that we only perform cross-attention on resolutions of $\frac{1}{8}$ and $\frac{1}{16}$.



\begin{table*}[t!]
	\caption{\textbf{Quantitative results with SOTA methods on benchmark datasets.} `$\uparrow$' and `$\downarrow$' mean higher and lower are desired. The Sync$_{conf}$ are marked in gray for its weak reflection of audio-visual synchronization. The runtime is evaluated on V$100$.}
	\vspace{-3mm}
	\label{tab:sota}
	\centering%
	\resizebox{0.97\linewidth}{!}{  
\begin{tabular}{@{}lcrrrrrrGllrGc@{}}
\toprule[1pt]
\multirow{2}{*}{Method} & \multicolumn{1}{c}{\multirow{2}{*}{Dataset}} & \multicolumn{7}{c}{Reconstruction}                                                                                                                                          & \multicolumn{1}{c}{} & \multicolumn{3}{c}{Cross Audio}             & \multicolumn{1}{c}{\multirow{2}{*}{\makecell[c]{Runtime\\(ms)}}}                                         \\ \cmidrule(l){3-9} \cmidrule(l){10-13}  
                        & \multicolumn{1}{c}{}                         & \multicolumn{1}{l}{LMD$\downarrow$} & 
                        \multicolumn{1}{l}{M-LMD$\downarrow$} &
                        \multicolumn{1}{l}{F-LMD$\downarrow$} &
                        \multicolumn{1}{l}{FID$\downarrow$} & \multicolumn{1}{l}{LPIPS$\downarrow$} & \multicolumn{1}{l}{SSIM$\uparrow$}  & \multicolumn{1}{l}{Sync$_{conf}$} &                      & FID$\downarrow$ & \multicolumn{1}{l}{CSIM$\uparrow$} & \multicolumn{1}{l}{Sync$_{conf}$}  \\  \bottomrule
GT                      & \multirow{7}{*}{VoxCeleb1~\cite{Nagrani17}}                   & \multicolumn{1}{c}{-} & \multicolumn{1}{c}{-} & \multicolumn{1}{c}{-}   & \multicolumn{1}{c}{-}   & \multicolumn{1}{c}{-}    & \multicolumn{1}{c}{-}                               & 6.54                          &                  &   \multicolumn{1}{c}{-}   & \multicolumn{1}{c}{-}                              & 6.54                  & \multicolumn{1}{c}{-}             \\ 
Wav2Lip~\cite{prajwal2020lip}                 &                                              & 8.78      & 15.87  & 5.82               & 17.58                                 & 0.1097         & 0.9351                                             & 7.23                          &        &   20.87                & 0.9329                              & 6.73            &        8.3       \\
PC-AVS ~\cite{zhou2021pose}                 &                                              & 23.90    & 22.08 & 24.66               & 65.21        & 0.3281             & 0.6960                                                                    & 7.17                          &      &      69.41               & 0.7621                              & 7.20         &          51.7        \\

TalkLip ~\cite{wang2023seeing}                &                                              & 19.29    & 31.83 & 14.06                & 34.70              & 0.1557             & 0.8987                                                             & 6.92                          &         &     22.27             & 0.9238                            & 4.90                   &17.4            \\
DINet~\cite{zhang2023dinet}                   &                                              & 18.25    & 24.57 & 15.61               & 23.83        & 0.1235              & 0.9091                                                                 & 5.52                          &       &      27.56              & 0.8385                              & 4.66           &     129.9          \\
IP-LAP~\cite{zhong2023identity}                 &                                              & 8.69   & 14.44 & 6.29                  & 16.84          & 0.1196            & 0.9279                                                                & 6.05                          &        &       23.81            & 0.9287                              & 4.20                   & 381.5          \\ \midrule
Ours                    &                                              & \textbf{6.72}   & \textbf{11.02} & \textbf{4.92}         & \textbf{12.73}       & \textbf{0.0916}     & \textbf{0.9361}                                     & 6.21                          &     &    \textbf{17.52}                  & \textbf{0.9434}                     & 5.00        & 33.1                          \\  \midrule[1pt]
GT                      & \multirow{7}{*}{MEAD~\cite{kaisiyuan2020mead}}                        & \multicolumn{1}{c}{-} & \multicolumn{1}{c}{-} & \multicolumn{1}{c}{-}  & \multicolumn{1}{c}{-}   & \multicolumn{1}{c}{-}    & \multicolumn{1}{c}{-}                                 & 4.65                          &           & \multicolumn{1}{c}{-}     & \multicolumn{1}{c}{-}                               & 4.65                       & \multicolumn{1}{c}{-}                 \\
Wav2Lip~\cite{prajwal2020lip}                 &                                              & 13.20      & 29.02 & 6.61              & 24.97          & 0.1346               & 0.9219                                                             & 6.87                          &          &     23.68            & 0.9307                             & 6.73                  &        8.3                       \\
PC-AVS ~\cite{zhou2021pose}                 &                                              & 20.82    & 26.73 & 18.35
                & 86.02            & 0.3457             & 0.7553                                                              & 7.26                          &      &     90.81                & 0.7550                             & 7.69                     &          51.7                       \\

TalkLip~\cite{wang2023seeing}                 &                                              & 16.80   & 34.10 & 9.59                 & 35.64        & 0.1622                & 0.9073                                                            & 6.75                          &     &     29.34                 & 0.9316                              & 4.94                 &17.4                         \\
DINet~\cite{zhang2023dinet}                   & & 15.33  & 38.14 & 5.82                  & 23.90                 & 0.1131         & 0.9236                                                            & 5.14                          &     &  24.16                    & 0.8099                              & 4.70                     &     129.9                   \\
IP-LAP~\cite{zhong2023identity}                 &                                              & 9.22    & 18.68 & 5.27               & 31.57              & 0.1441         & \textbf{0.9285}                                                               & 5.77                          &    &      36.68                 & 0.9472                             & 4.01                         & 381.5               \\ \midrule
Ours                    &                                              & \textbf{9.04}   & \textbf{18.65} & \textbf{5.02}         & \textbf{11.68}       & \textbf{0.0958}        & 0.9251                                  & 4.00                             &       &      \textbf{13.22}              & \textbf{0.9638}                     & 3.84         & 33.1                                \\ \midrule[1pt]

DINet~\cite{zhang2023dinet}    & \multirow{3}{*}{HDTF~\cite{2021Flow}} & 8.0255 & 15.36        & 5.185          & 12.94  & 0.0975   & 0.9327                                  & \multicolumn{1}{c}{-}                           &           & \multicolumn{1}{c}{-}     & \multicolumn{1}{c}{-}                               & \multicolumn{1}{c}{-}                          & 129.9                    \\
IP-LAP~\cite{zhong2023identity}  & & 6.076  & 10.20        & 4.658          & 9.490     &   0.1101 &   0.9416                                                            & \multicolumn{1}{c}{-}                         &          &     \multicolumn{1}{c}{-}          & \multicolumn{1}{c}{-}                           & \multicolumn{1}{c}{-}               &        381.5                      \\ 
Ours  & & \textbf{6.011}  & \textbf{9.966} & \textbf{4.207} & \textbf{6.065} &  \textbf{0.0820}  &  \textbf{0.9418}                                 & \multicolumn{1}{c}{-}                              &       &     \multicolumn{1}{c}{-}              & \multicolumn{1}{c}{-}                  & \multicolumn{1}{c}{-}         & 33.1                                \\ \bottomrule
\end{tabular}

	}
	\vspace{-3mm}
\end{table*}

\begin{figure*}[t!]
    \centering
    \includegraphics[width=0.99\linewidth]{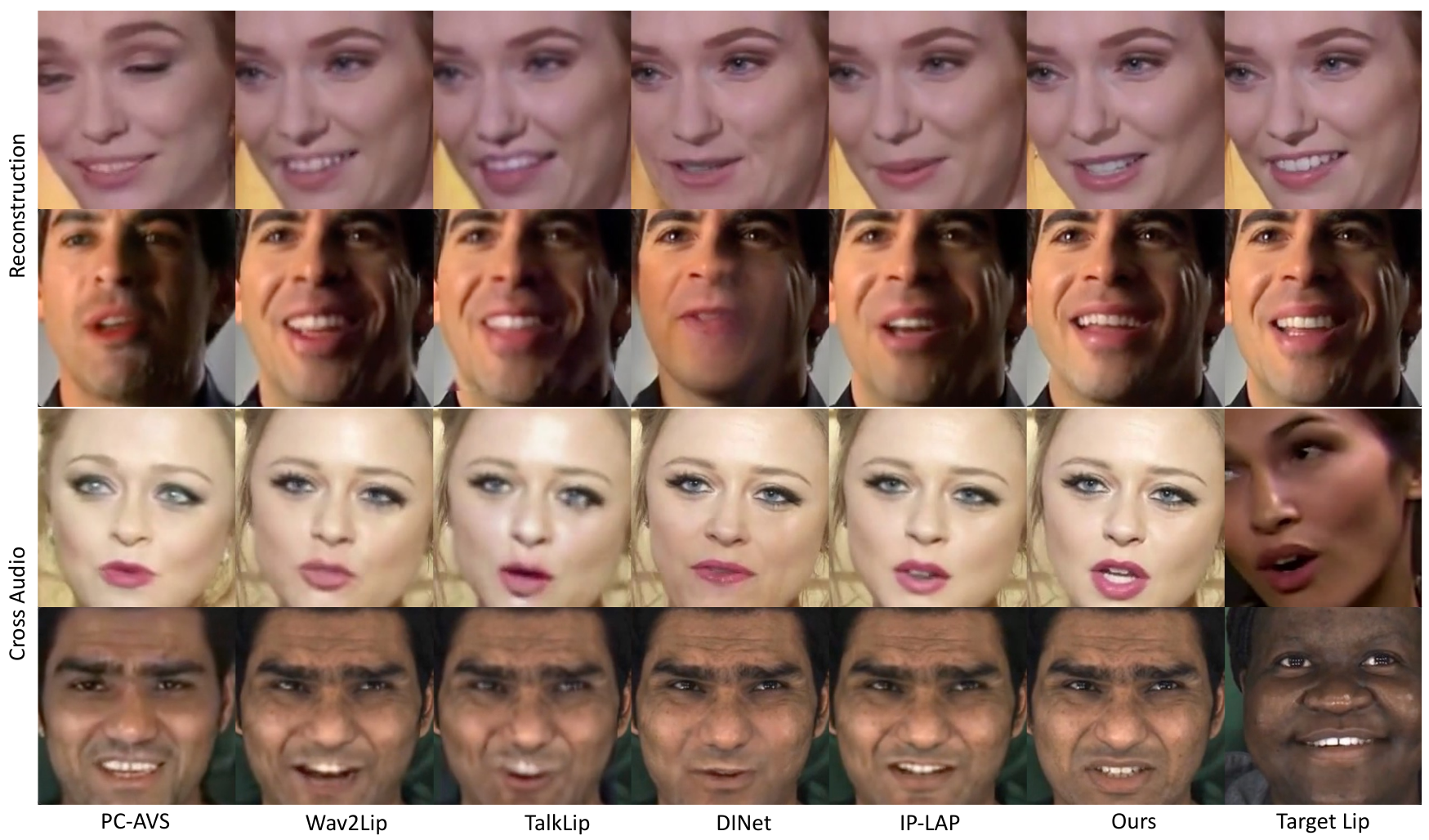} 
    \vspace{-3mm}
    \caption{\textbf{Visual comparisons with state-of-the-art competitors.} Our method achieves the best lip-speech sync and visual quality.}
    \label{fig:sota}
\vspace{-2mm}

\end{figure*}
\vspace{-3mm}

\section{Experiments}

\subsection{Experiments Setting}

\noindent \textbf{Implementation Details.} 
In our experiments, $N=16$, $T=16$, $l=32$. The $N$ historical expressions are randomly selected during training and inference.  We train the expression-to-face renderer at $256\times256$ resolution. The identity reference is set to the first frame of a video clip. The number of encoder and decoder stages $d$ is $4$, and each stage has $2$ stacked residual blocks. More details are included in the supplementary material.

\noindent \textbf{Datasets.} 
We conducted experiments on three popular datasets: VoxCeleb$1$~\cite{Nagrani17}, MEAD~\cite{kaisiyuan2020mead}, HDTF~\cite{2021Flow}. VoxCeleb$1$ comprises over $100,000$ utterances from $1,251$ celebrities, extracted from videos uploaded to YouTube. 
We utilized these utterances that are available from Internet (about  $10$\% of total) for training and randomly selected $50$ utterances for evaluation. 
MEAD is a talking-face video corpus that features $60$ actors and actresses expressing $8$ different emotions at $3$ distinct intensity levels. 
We exclusively use the front view videos, selecting $40$ actors for training and $3$ actors for evaluation. 
20 video clips in HDTF testset are used only for evaluation under reconstruction setting.

\noindent \textbf{Evaluation Metrics.}
We employ facial Landmarks Distance (LMD)~\cite{bulat2017far} to assess the accuracy of lip-sync, M- (Mouth) and F- (Face) separately for better evaluation, FID (Fréchet Inception Distance)~\cite{heusel2017gans}, LPIPS (Learned Perceptual Image Patch Similarity)~\cite{zhang2018unreasonable}, and SSIM (Structural Similarity Index Measure) to evaluate the quality of generated images. 
We also assess the identity preservation of generated faces by measuring  CSIM (cosine similarity between identity features)~\cite{schroff2015facenet}. 
LMD assesses lip-audio sync accuracy among different methods, with a lower value indicating closer alignment with GT and thus better synchronization with the audio.
A lower FID signifies that the generated image quality more closely resembles the original video, reflecting image clarity and naturalness.
Moreover, a higher CSIM indicates higher facial similarity, suggesting superior identity preservation by the corresponding method. 
Additionally, we employ the SyncNet~\cite{chung2017out} metric to evaluate audio-visual consistency. 
However, it is crucial to emphasize that a higher SyncNet score doesn't necessarily indicate better audio-visual sync, as discussed in~\cite{guan2023stylesync}.


\begin{table}[t]
\begin{minipage}[t]{0.49\linewidth}
\centering
\makeatletter\def\@captype{table}\makeatother\caption{Mean Opinion Score (MOS) on benchmark datasets.}
\label{tab:mos}
\vspace{-0.2cm}
	\resizebox{0.99\linewidth}{!}{
	\begin{tabular}{lccccc}
\toprule
\makecell[c]{MOS / Method}   & \multicolumn{1}{c}{Wav2Lip} & \multicolumn{1}{c}{DINet} & \multicolumn{1}{c}{TalkLip} & \multicolumn{1}{c}{IP-LAP} & \multicolumn{1}{c}{Ours} \\ \hline
Visual Quality & 1.53                        & 1.72                      & 1.55                        & 2.84                       & \textbf{3.77}  (33\%$\uparrow$)           \\
Lip Sync       & 2.15                        & 2.50                       & 2.06                        & 2.58                       & \textbf{3.72}      (44\%$\uparrow$)       \\ \bottomrule
\end{tabular}
	}
\end{minipage}
\begin{minipage}[t]{0.49\linewidth}
\centering
\makeatletter\def\@captype{table}\makeatother\caption{Metric and runtime comparison with StyleTalk.}
\label{tab:styletalk}
\vspace{-0.2cm}
	\resizebox{0.99\linewidth}{!}{
\begin{tabular}{@{}lcccccc@{}}
\toprule
Metric    & M-LMD  $\downarrow$         & F-LMD  $\downarrow$         & FID   $\downarrow$  & LPIPS  $\downarrow$ & SSIM  $\uparrow$         & Runtime (ms) $\downarrow$ \\ \midrule
StyleTalk~\cite{ma2023styletalk} & 7.910          & 5.220          & 15.42     & 0.1486 &  0.8016   & 141.0           \\
Ours      & \textbf{4.387} & \textbf{2.215} & \textbf{12.25} & \textbf{0.0908}&  \textbf{0.9093}& \textbf{33.1} \\ \bottomrule
\end{tabular}
	}
\end{minipage}
\end{table}
\vspace{-0.2cm}

 

\begin{figure}[t]
\begin{minipage}[t]{0.49\linewidth}
\centering
\vspace{-0.2cm}
	\resizebox{0.99\linewidth}{!}{
	\includegraphics[width=0.9\linewidth]{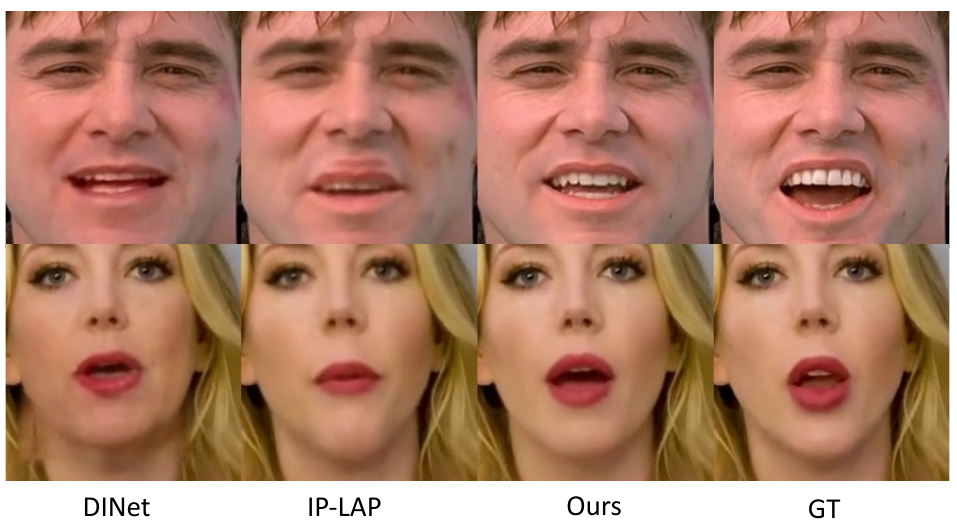}
	}
\vspace{-0.5cm}
\makeatletter\def\@captype{figure}\makeatother\caption{Visual results on HDTF. Please zoom in for more detail.}
\label{fig:hdtf}
\end{minipage}
\begin{minipage}[t]{0.49\linewidth}
\centering
\vspace{-0.2cm}
	\resizebox{0.74\linewidth}{!}{
	\includegraphics[width=0.9\linewidth]{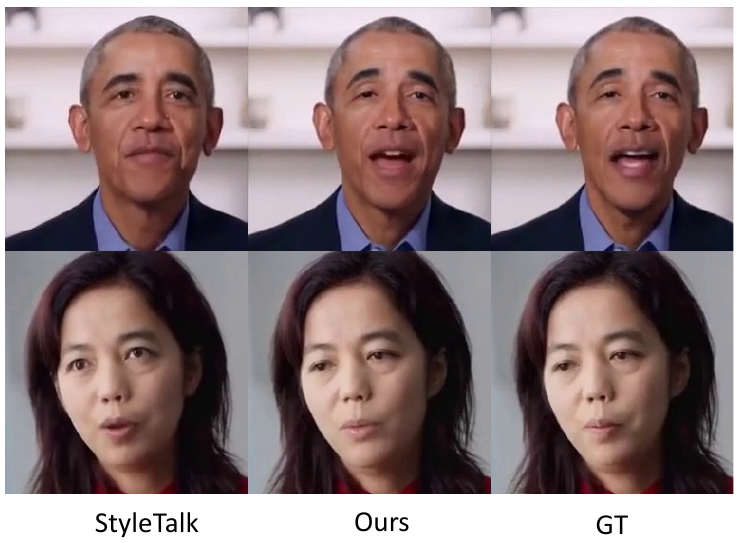}
	}
\vspace{-0.2cm}
 \makeatletter\def\@captype{figure}\makeatother\caption{Visual comparison with StyleTalk on its official demos.}
\label{fig:styletalk}
\end{minipage}
\vspace{-0.3cm}
\end{figure}
\vspace{-2mm}

\subsection{Comparison with SOTA Methods}
\vspace{-2mm}

\noindent \textbf{Comparison Methods.}
We compare the proposed method with state-of-the-art person-generic audio-driven face generation methods, including Wav2Lip~\cite{prajwal2020lip}, PC-AVS~\cite{zhou2021pose}, DINet~\cite{zhang2023dinet}, TalkLip~\cite{wang2023seeing}, and IP-LAP~\cite{zhong2023identity}. 
Among these methods, Wav2Lip excels in reconstruction performance.  
PC-AVS stands out for editing lip motions and poses. 
DINet employs deformable convolution to construct its reconstruction network. 
TalkLip, structurally similar to Wav2Lip, introduces a new lip-reading loss function. 
IP-LAP utilizes $2$D landmarks as intermediate information, serving as our primary comparison method.

\noindent \textbf{Reconstruction.} 
In this setting, the video clip is driven by the corresponding original audio. 
Since the video clip illustrates the target lip motions, it serves as the ground truth for calculating metrics requiring paired data. 

As shown in Table~\ref{tab:sota}, our method achieves the best results among most metrics on all three datasets.
Our method exhibits a significant advantage in the FID metric, surpassing the second-best method by  $51$\% and $36\%$ on MEAD and HDTF, respectively. 
IP-LAP, employing multiple reference frames (\textit{i.e.}, $25$ frames), achieves comparable results on LMD but severely impacts efficiency (\textit{i.e.}, $10$$\times$ slower than~\name). 
Although Wav2Lip achieves noteworthy outcomes on metrics, our examination reveals its inadequate visual quality, as corroborated by the user study results in Table~\ref{tab:mos}. \par


The top two rows of Fig.~\ref{fig:sota} and~\ref{fig:hdtf} display the qualitative results under the reconstruction setting, with the rightmost column representing the ground truth. 
By incorporating the improved facial prior, our proposed audio-to-expression transformer can precisely predict lip shapes according to the individual's movement amplitude, resulting in outcomes closer to the GT. 
Furthermore, our method captures similar textures resembling the original face with a single image, demonstrating the efficacy and efficiency of our FIA module.
Concerning the SyncNet metric, although both Wav2Lip and PC-AVS scored exceptionally high, the audio-visual synchronization did not improve correspondingly, creating inconsistency with the visual effects. 

\noindent \textbf{Dubbing with Cross Audio.}
In this setting, the video clip is driven by another audio segment, providing a more representative depiction of real-world scenarios.
Our method achieves the best FID on both datasets, indicating more natural and realistic results in cross-audio settings. 
The bottom two rows of Fig.~\ref{fig:sota} depict qualitative results under cross-audio testing.
The rightmost column displays the lip motion from the video aligned to the cross audio, serving as a pseudo GT for the accurate lip shape. 
Our results exhibit closer lip shapes to the pseudo GT, indicating better lip-speech sync against the SOTA competitors.

We further conduct a user study to evaluate the generation quality and lip synchronization of different methods. 
$10$ videos were selected, and scores were collected from $15$ participants, ranging from $1$ (worst) to $5$ (best). 
As shown in Table~\ref{tab:mos}, our method excels in both generation quality and lip synchronization, outperforming the second-best IP-LAP by a significant margin (\ie, $33$\% and $44$\% improvements).

\noindent \textbf{Runtime Analysis.}
Our approach outperforms the SOTA methods, speeding up $3.92$$\times$ and $11.5$$\times$ faster than DINet and IP-LAP, respectively. Specifically, the runtime of the audio-to-expression transformer and the expression-to-face renderer are $2.3$ms and $30.8$ms, respectively.
By utilizing $3$D priors, our method facilitates the precise expression generation with reduced computational requirements.
Our FIA module efficiently executes reference texture transfer, thereby eliminating the necessity for multi-encoder and multi-reference feature alignment processes. 
Moreover, our method offers a distinct quality advantage over the real-time methods (e.g., TalkLip, Wav2Lip). 
As evidenced by the reconstruction metrics in Table~\ref{tab:sota}, despite its rapid processing speed, TalkLip's reconstruction capability is subpar. 
Visual inspection reveals that Wav2Lip, the fastest method, presents significant artifacts in cross-audio scenarios, leading to a decrease in generalization performance. 
Overall, our method constitutes the optimal solution for achieving a balance between effectiveness and efficiency.


\noindent \textbf{Comparison with One-shot Method.}
We emphasize the main differentials compared with one-shot talking head methods, \textit{e.g.}, StyleTalk~\cite{ma2023styletalk}. In \name, we explicitly introduce historical expressions and focus on local lip movements by vertex loss in Equation~($4$). 
In StyleTalk, the style is implicit and is used to control global head and facial movements. 
As a lip movement specialist, \name~exhibits superior lip-synchronization than StyleTalk in comparison on their official demos, as shown in Tab.~\ref{tab:styletalk} and Fig.~\ref{fig:styletalk}. 
In summary, \name~outperforms in all $5$ metrics and is $4.27$$\times$ faster than StyleTalk, which adopts PIRender~\cite{ren2021pirenderer} in the second stage.

\vspace{-0.2cm}

\begin{table}[t]
\begin{minipage}[t]{0.49\linewidth}
\centering
\makeatletter\def\@captype{table}\makeatother\caption{Ablation study on different facial priors.}
\label{tab:abl_prior}
\vspace{-0.2cm}
	\resizebox{0.99\linewidth}{!}{

\begin{tabular}{@{}llcccc@{}}
\toprule
\multirow{2}{*}{\makecell[c]{Facial\\Prior}} & \multicolumn{1}{c}{Shape}                & $\times$                  & $\checkmark$               & $\times$                   & $\checkmark$               \\
                              & \multicolumn{1}{l}{\makecell[c]{Historical Expression}} & $\times$                  & $\times$                   & $\checkmark$               & $\checkmark$               \\ \midrule
\multicolumn{2}{l}{\makecell[l]{Expression Error}}                 & \multicolumn{1}{r}{0.2680} & \multicolumn{1}{r}{0.1342} & \multicolumn{1}{r}{0.1186} & \multicolumn{1}{r}{\bf{0.1128}} \\ \bottomrule
\end{tabular}
	}
\end{minipage}
\begin{minipage}[t]{0.49\linewidth}
\centering
\makeatletter\def\@captype{table}\makeatother\caption{Reconstructions with different masks on VoxCeleb1.}
\label{tab:abl_mask}
\vspace{-0.2cm}
	\resizebox{0.75\linewidth}{!}{
	\begin{tabular}{@{}lllll@{}}
\toprule
Mask      & LMD$\downarrow$                              & FID   $\downarrow$                                & LPIPS    $\downarrow$                              & SSIM     $\uparrow$                              \\ \midrule
Naive     &   \multicolumn{1}{r}{7.08}                                &      \multicolumn{1}{r}{13.31}                              &        \multicolumn{1}{r}{0.1133}                             &   \multicolumn{1}{r}{0.9088}                                  \\
Learnable & \multicolumn{1}{r}{\textbf{6.72}} & \multicolumn{1}{r}{\textbf{12.73}} & \multicolumn{1}{r}{\textbf{0.0916}} & \multicolumn{1}{r}{\textbf{0.9361}} \\ \bottomrule
\end{tabular}
	}
\end{minipage}
\vspace{-0.2cm}
\end{table}

\subsection{Ablation Study}

\begin{figure}[t!]
	\centering
	\includegraphics[width=0.8\linewidth]{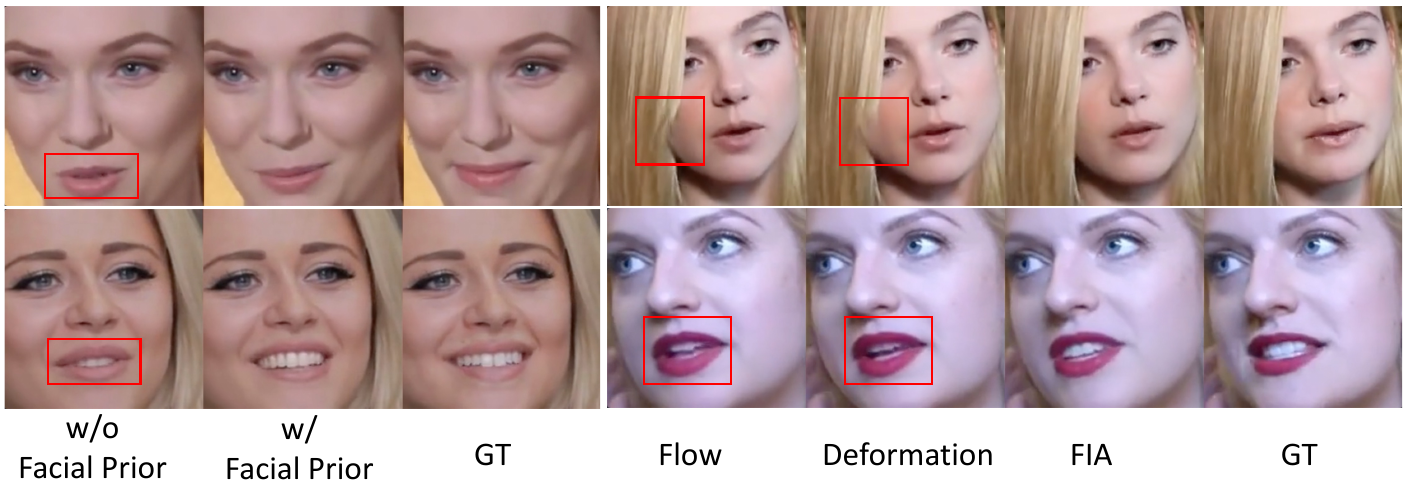} 
	\vspace{-2mm}
	\caption{\textbf{Left}: Effects on with or without improved facial priors. 
		\textbf{Right}: Effects on FIA paired with different reference module.}
	\label{fig:abl}
\vspace{-3mm}
 
\end{figure}

\noindent \textbf{Effectiveness of the Improved Facial Prior.}
To validate the effectiveness of our improved facial priors, we separately removed the shape and historical expressions, and evaluate on VoxCeleb1 dataset under reconstruction setting. 
Since the GT expression coefficients are known, we directly quantified the mean squared error between the predicted expression coefficients of different models and the GT coefficients. 
In Table~\ref{tab:abl_prior}, introducing both shape and historical expressions positively impacts lip synchronization prediction.
Compared to the model without shape and expression prior, the full model improves prediction accuracy by $57.9$\%. 
The visual results depicted in Fig.~\ref{fig:abl}-Left demonstrate the efficacy of the enhanced facial prior in maintaining intra-personal expressions.


\begin{table}[t!]
	\caption{Reconstruction performance of FIA with different reference module and across different scales on VoxCeleb$1$. The runtime only reflects the face renderer.}
	\vspace{-3mm}
	\label{tab:abl_attn}
	\centering%
	\resizebox{0.8\linewidth}{!}{  
	\setlength{\tabcolsep}{12pt}
\begin{tabular}{@{}lccccc@{}}
\toprule
Configuration    & \multicolumn{1}{c}{FID$\downarrow$} & \multicolumn{1}{c}{LPIPS$\downarrow$} & \multicolumn{1}{c}{SSIM$\uparrow$}   & \multicolumn{1}{c}{\makecell[c]{Runtime\\(ms)}} & \multicolumn{1}{c}{\makecell[c]{Param.\\(M)}} \\ \midrule
Flow      & 13.68                   & 0.0963                    & 0.9329               & 30.48                   & 82.94                \\
Deformation    & 13.38                   & 0.0948                    & 0.9332               &    31.20                    & 98.79                 \\

Blocks=1&         13.42          &     0.0959   &   0.9326     & 24.28    &      52.53    \\

Blocks=3&           \textbf{12.11}        &      0.0924     &     0.9352          & 38.65    &          85.96       \\

Final (FIA) & 12.73                 & \textbf{0.0916}           & \textbf{0.9361}            &    30.82            & 69.24                 \\

\bottomrule

\end{tabular}
	}
	\vspace{-3mm}
\end{table}

\noindent \textbf{Effectiveness of the Learnable Mask.}
To validate the benefits of our proposed learnable mask in talking face generation, we replace it with a naive mask obscuring the lower half of the image and conduct experiment on VoxCeleb1 dataset under reconstruction setting. 
As shown in Table~\ref{tab:abl_mask}, the performance drops when using the naive mask. 
The naive mask lacks information about the target face shape and includes irrelevant background in the area that the network has to generate, posing increased learning difficulty. 
Conversely, the learnable mask is intrinsically associated with the target audio, remaining impervious to the original facial contour, thereby guaranteeing enhanced accuracy of lip movements and the naturalness of facial expressions.


\noindent \textbf{Comparison with Deformation and Flow-based Module.}
To validate our FIA module, we replace its cross-attention component with other common alignment modules: flow-based warping and deformation convolution. Table~\ref{tab:abl_attn} shows the impact of building FIA with different modules on generated image quality, along with comparisons in terms of runtime and parameters. 
FIA (Final in Table~\ref{tab:abl_attn}) achieves superior visual quality with fewer parameters over the deformation and flow-based structure.
Fig.~\ref{fig:abl}-Right illustrates the proposed FIA paired with cross-attention, highlighting its strong ability to restore textures, such as hair and teeth. Unlike flow-based and deformable convolution methods overly relying on the reference, cross-attention allows for a weighted fusion of features from different regions, enabling a more flexible generation of facial textures.


\noindent \textbf{Complexity.}
Table~\ref{tab:abl_attn} also displays the impact of adjusting the number of residual blocks within the FIA module on generated image quality, assessing the performance of~\name~across different scales.
Configurations with $1$, $2$ (final model), and $3$ residual blocks are explored. 
Setting num=$1$ accelerates speed but diminishes visual quality, while num=$3$ improves FID results at the expense of real-time performance. 
Consequently, num=$2$ strikes a good balance between visual quality and speed.
\vspace{-0.3cm}

\section{Conclusion}
\vspace{-2mm}


We propose a novel audio-driven framework \name, incorporating an audio-to-expression transformer and a high-fidelity expression-to-face renderer. Our improved facial prior adeptly adjusts speech content while maintaining identity through a cross-modal attention on both identity and intra-persona variation features. A specialized learnable mask tackles challenges associated with altering facial structures. Our FIA module, combining AdaIN and cross-attention structures, facilitates precise lip-shape control using 3D coefficients and efficient facial texture transfer from a single frame. Experimental results on benchmarks affirm our method’s superiority in lip-speech sync and generation quality, emphasizing its efficiency and applicability.\par

\noindent \textbf{Limitation and Social Impacts.}
Our approach encounters limitations with facial obstructions, such as microphones or hand movements in front of the face. 
This is expected, given our primary focus on facial generation, particularly in modeling mouth shapes, without an additional segmentation model to predict facial obstructions.
Efficient talking face technology can be used for digital human live streaming and interaction, but it also carries risks in illicit industries, including the manipulation of spoken content for deceptive purposes. 
To prevent misuse, generated videos should be clearly marked. 
Ongoing research should also be dedicated to identifying AI-generated videos, evolving alongside advancements in generative models. \par

\bibliographystyle{splncs04}
\bibliography{main}
\end{document}


\newcommand{\name}{$\mathtt{RealTalk}$}

\title{RealTalk: Real-time and Realistic Audio-driven Face Generation with 3D Facial Prior-guided Identity Alignment Network} 


\author{First Author\inst{1}\orcidlink{0000-1111-2222-3333} \and
Second Author\inst{2,3}\orcidlink{1111-2222-3333-4444} \and
Third Author\inst{3}\orcidlink{2222--3333-4444-5555}}

\authorrunning{F.~Author et al.}

\institute{Princeton University, Princeton NJ 08544, USA \and
Springer Heidelberg, Tiergartenstr.~17, 69121 Heidelberg, Germany
\email{lncs@springer.com}\\
\url{http://www.springer.com/gp/computer-science/lncs} \and
ABC Institute, Rupert-Karls-University Heidelberg, Heidelberg, Germany\\
\email{\{abc,lncs\}@uni-heidelberg.de}}

\maketitle


\setcounter{page}{1}

\section{Training Details}
In this section, we elucidate the specifics and configurations employed during the training process. 

In the preprocessing phase, videos are sampled at a rate of $25$ frames per second (FPS). Frames with faces undergo face detection to yield face bounding boxes, followed by the D$3$DFR~\cite{blanz1999morphable} model predicting $3$D coefficients. The D$3$DFR model, initially fine-tuned on VoxCeleb1~\cite{Nagrani17} and MEAD~\cite{kaisiyuan2020mead}, is utilized to enhance reconstruction precision. Audio features are extracted via the Hubert~\cite{hsu2021hubert} large model pretrained on 16kHz sampled audio with output feature at a sampling rate of $50$ FPS.

In the audio-to-expression phase, the encoder and decoder comprise $8$ layers, each with a multi-head attention layer with $8$ heads and latent dimension of $256$. The training was executed on $8$ NVIDIA V$100$, with a batch size of $64\times8$ and a learning rate of $5$e$-5$. The Adam optimizer was employed, with $\beta_1$ and $\beta_2$ values of $0.95$ and $0.999$ respectively, over $200$k iterations.

In the expression-to-face phase, the training batch size was $10\times8$, with a learning rate of $1$e$-4$. The Adam optimizer was used, with $\beta_1$ and $\beta_2$ values of $0.9$ and $0.96$ respectively, a weight decay of $1$e$-5$, and $100$k iterations.

\paragraph{Architecture of the Expression-to-Face Renderer.}
The comprehensive network structure of the expression-to-face renderer is delineated in Table~\ref{tab:arc_e2f}. The network is formed in an encoder-decoder style, where the input image is progressively downsampled and upsampled. Note that, we only perform cross attention operation on Dec. L$0$ and Dec. L$1$ as we find that attention operation on larger resolution brings less visual quality improvement but significantly higher computational effort.
\begin{table}[t!]
	\caption{\textbf{Details of the expression-to-face renderer architecture.} `Enc.': Encoder, `Dec.': Decoder, `Conv.': Convolution, `K.': Kernel size, `S.': Stride, `BN.': Batch Normalization, `Res.': Residual, `Attn.': Attention, `Interp.': Interpolation.}
	\centering%
	\begin{tabular}{c|c|c} 
\toprule
\begin{tabular}[c]{@{}c@{}}\\\\\\\\\end{tabular} & Output Size                                  & Operation                                                                                              \\ 
\hline\hline
\multirow{2}{*}{Enc. L0}                          & \multirow{2}{*}{$256^2 \times 32$}  & Input Conv. K.$3\times3$, S.1                                                                                     \\
                                                 &                                              & $[\frac{\operatorname{BN. Conv. K.3\times3, S.1} \times 2}{\operatorname{ReLU + Res.}}] \times 2 $    \\ 
\midrule
\multirow{2}{*}{Enc. L1}                          & \multirow{2}{*}{$128^2 \times 64$} & Down Conv. K.$3\times3$, S.2                                                                                     \\
                                                 &                                              & $[\frac{\operatorname{BN. Conv. K.3\times3, S.1} \times 2}{\operatorname{ReLU + Res.}}] \times 2 $   \\ 
\midrule
\multirow{2}{*}{Enc. L2}                          & \multirow{2}{*}{$64^2 \times 128$} & Down Conv. K.$3\times3$, S.2                                                                                   \\
                                                 &                                              & $[\frac{\operatorname{BN. Conv. K.3\times3, S.1} \times 2}{\operatorname{ReLU + Res.}}] \times 2 $  \\ 
\midrule
\multirow{2}{*}{Enc. L3}                          & \multirow{2}{*}{$32^2 \times 256$}   & Down Conv. K.$3\times3$, S.2                                                                                   \\
                                                 &                                              & $[\frac{\operatorname{BN. Conv. K.3\times3, S.1} \times 2}{\operatorname{ReLU + Res.}}] \times 2 $  \\ 
\midrule
\multirow{2}{*}{Enc. L4}                          & \multirow{2}{*}{$16^2 \times512$}   &     Down Conv. K.$3\times3$, S.2                                                                               \\
                                                 &                                              & $[\frac{\operatorname{BN. Conv. K.3\times3, S.1} \times 2}{\operatorname{ReLU + Res.}}] \times 2 $                                                                            \\ 
\midrule
\multirow{3}{*}{Dec. L0}                          & \multirow{3}{*}{$16^2 \times512$}   & Input Conv. K.$3\times3$, S.1                                                                                      \\
                                                 &                                              & $[\frac{\operatorname{AdaIN. Conv. K.3\times3, S.1} \times 2}{\operatorname{ReLU + Res.}}] \times 2 $    \\
                                                 &
                                                 & Cross Attn. with Enc. L4
                                                 \\ 
\midrule
\multirow{3}{*}{Dec. L1}                          & \multirow{3}{*}{$32^2 \times256$}   & Interp. Up Conv. K.$3\times3$, S.1                                                                                    \\
                                                 &                                              & $[\frac{\operatorname{AdaIN. Conv. K.3\times3, S.1} \times 2}{\operatorname{ReLU + Res.}}] \times 2 $                                                                              \\ 
                                                  &
                                                 & Cross Attn. with Enc. L3\\
\midrule
\multirow{2}{*}{Dec. L2}                          & \multirow{2}{*}{$64^2\times128$}   & Interp. Up Conv. K.$3\times3$, S.1                                                                                    \\
                                                 &                                              & $[\frac{\operatorname{AdaIN. Conv. K.3\times3, S.1} \times 2}{\operatorname{ReLU + Res.}}] \times 2 $                                                                              \\ 
\midrule
\multirow{2}{*}{Dec. L3}                          & \multirow{2}{*}{$128^2\times64$}   & Interp. Up Conv. K.$3\times3$, S.1                                                                                    \\
                                                 &                                              & $[\frac{\operatorname{AdaIN. Conv. K.3\times3, S.1} \times 2}{\operatorname{ReLU + Res.}}] \times 2 $                                                                              \\  
\midrule
\multirow{3}{*}{Dec. L4}                          & \multirow{3}{*}{$256^2\times3$}   & Interp. Up Conv. K.$3\times3$, S.1                                                                                    \\
                                                 &                                              &

                                                 $[\frac{\operatorname{AdaIN. Conv. K.3\times3, S.1} \times 2}{\operatorname{ReLU + Res.}}] \times 2 $   
                                                 \\
                                                 &
                                                 &
                                                 Output Conv. K.$3\times3$, S.1
                                                 \\      
                                                 \bottomrule                           
\end{tabular}
	\label{tab:arc_e2f}
\end{table}

\begin{figure*}[t!]
    \centering
    \includegraphics[width=0.99\linewidth]{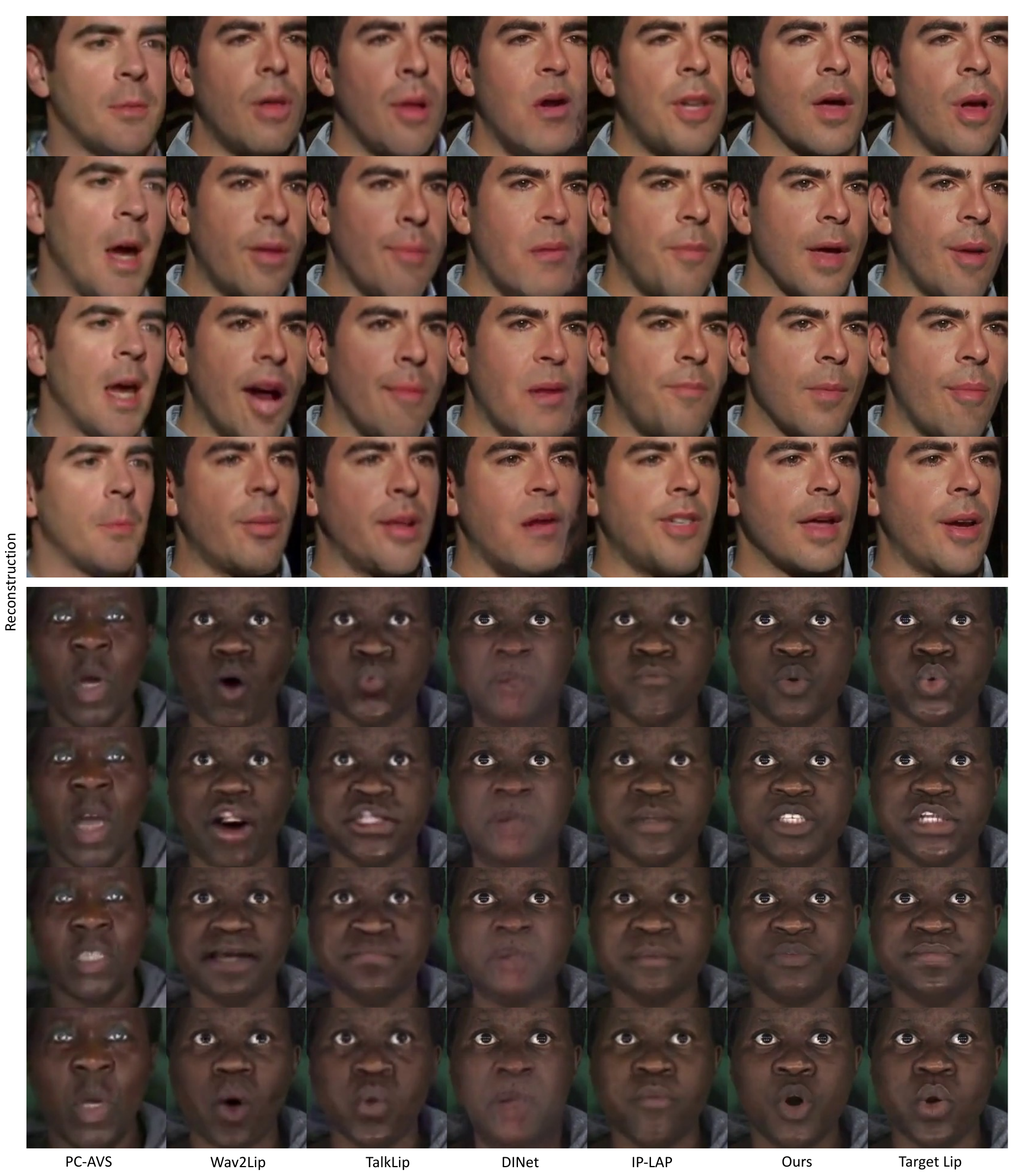} 
    \caption{Visual comparisons with state-of-the-art competitors under reconstruction setting.}
    \label{fig:X_sota_suppl_1}
\end{figure*}

\begin{figure*}[t!]
    \centering
    \includegraphics[width=0.99\linewidth]{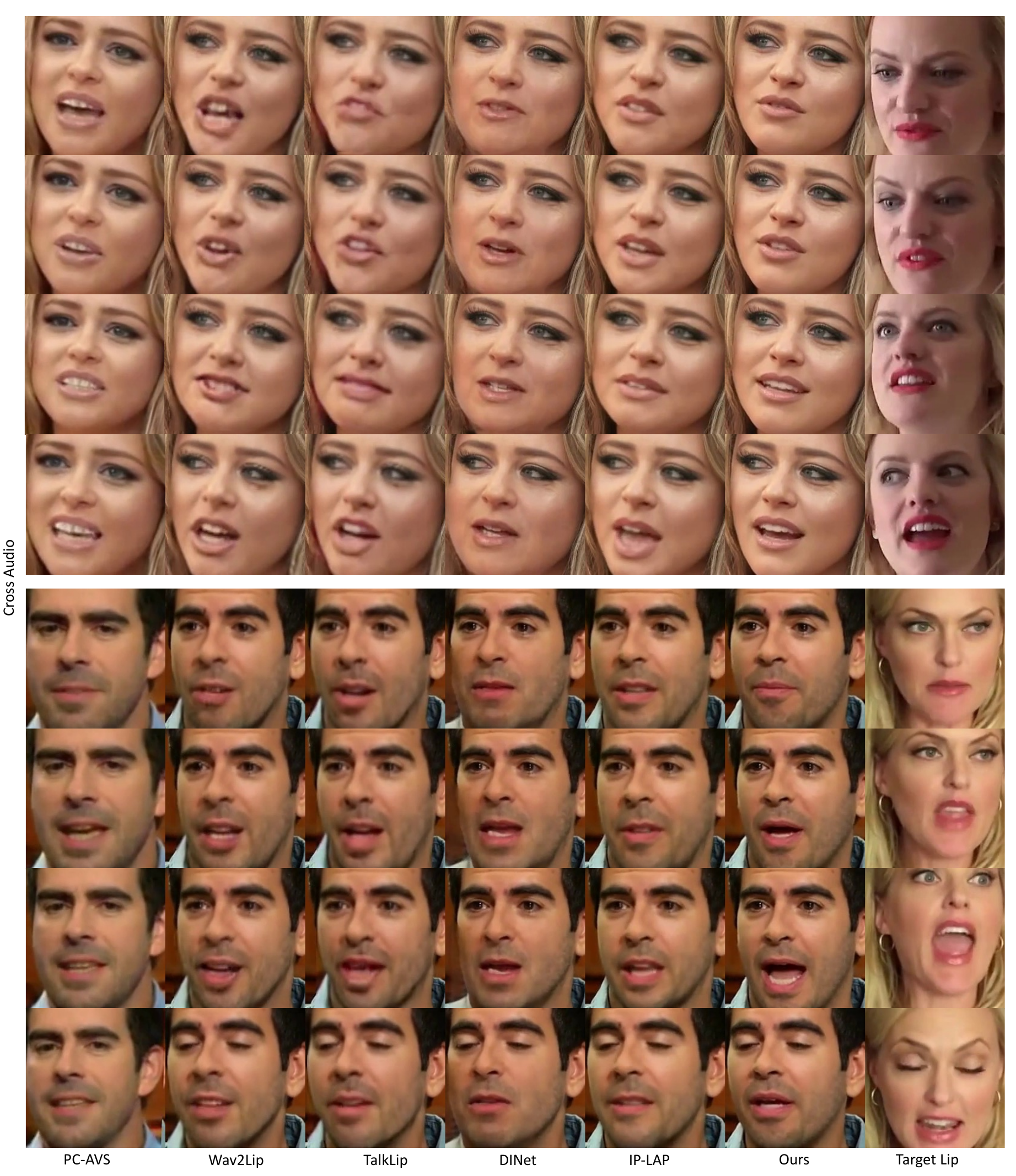} 
    \caption{Visual comparisons with state-of-the-art competitors under cross audio setting.}
    \label{fig:X_sota_suppl_2}
\end{figure*}

\section{Complementary Qualitative Comparison}
In this section, we offer an expanded visual comparison in Figure~\ref{fig:X_sota_suppl_1} and ~\ref{fig:X_sota_suppl_2}. The comparison methods include Wav2Lip~\cite{prajwal2020lip}, PC-AVS~\cite{zhou2021pose}, DINet~\cite{zhang2023dinet}, TalkLip~\cite{wang2023seeing}, and IP-LAP~\cite{zhong2023identity}. Our results achieve better lip sync according to the ``Target Lip'' both in reconstruction and cross audio setting. Moreover, our method generates faces with high visual quality.

\paragraph{Demo Video.}
Additionally, for a more comprehensive demonstration, a demo video (\href{run:RealTalk-Demo.mp4}{RealTalk-Demo.mp4})  is included in the supplementary material. We couple our RealTalk with Text-to-Speech technology and provide interesting interaction with celebrities. The upcoming videos indicate that the intermediate $3$D coefficients predicted by our RealTalk can describe lip movements precisely. Besides, our method can be generalized to multiple language, such as Chinese, Japanese, and Korean.

\bibliographystyle{splncs04}
\bibliography{main}